\documentclass{article}

\usepackage{arxiv}

\usepackage[utf8]{inputenc} 
\usepackage[T1]{fontenc}    
\usepackage{hyperref}       
\usepackage{url}            
\usepackage{booktabs}       
\usepackage{amsfonts}       
\usepackage{nicefrac}       
\usepackage{microtype}      
\usepackage{lipsum}
\usepackage{graphicx}
\usepackage{amsmath}
\usepackage[numbers]{natbib}

\graphicspath{ {./images/} }
\newcommand\E{\mathbb{E}}

\newcommand\ctx{\text{ctx}}
\newcommand\tar{\text{tgt}}

\title{Robust Neural Processes for Noisy Data}

\author{%
  Chen Shapira \\
  Department of Computer Science\\
  University of Haifa\\
  Israel \\
  \texttt{chen2600@gmail.com} \\
  \And
  Dan Rosenbaum \\
  Department of Computer Science \\
  University of Haifa \\
  Israel \\
  \texttt{danro@cs.haifa.ac.il} \\
}

\begin{document}
\date{}
\maketitle
\begin{abstract}
Models that adapt their predictions based on some given contexts, also known as in-context learning, have become ubiquitous in recent years. We propose to study the behavior of such models when data is contaminated by noise. Towards this goal we use the Neural Processes (NP) framework, as a simple and rigorous way to learn a distribution over functions, where predictions are based on a set of context  points. Using this framework, we find that the models that perform best on clean data, are different than the models that perform best on noisy data. Specifically, models that process the context using attention, are more severely affected by noise, leading to \textit{in-context overfitting}. We propose a simple method to train NP models that makes them more robust to noisy data. Experiments on 1D functions and 2D image datasets demonstrate that our method leads to models that outperform all other NP models for all noise levels.
\end{abstract}

\section{Introduction}
Recent years have seen blossoming research and applications of models that make predictions based on a given context. These models are used in different machine learning setups, including few shot learning~\cite{lakeHumanlevelConceptLearning2015, vinyals2017matching}, meta learning~\cite{santoro2016meta, finnModelAgnosticMetaLearningFast2017}, language modeling~\cite{brown2020language} and generative modelling in various domains~\cite{madani2023large}. In some cases,  such models exhibit impressive abilities to adapt their prediction based on the context, a phenomenon also called \emph{in-context learning}~\cite{chan2022data}.

We propose to study the performance of such models when data is noisy. Specifically, we test the performance when the context is contaminated with different levels of noise, and focus on the problem where training data also contains the same level of noise.
Towards this goal, we use the framework of neural processes~\cite{garneloConditionalNeuralProcesses2018, garneloNeuralProcesses2018}, a family of models that aim to capture probabilistic priors of data represented as functions. Under this representation, each data sample is a function that can be queried at different points, and the models are trained to predict the distribution of target points, given a set of observed context points. We argue that this framework provides an elegant and rigorous test-bed to study the fundamental problem of modeling functions under noisy conditions.

To demonstrate the problem of modeling functions with noisy data, we first define different possible setups, each using different assumptions on the noise and the access to clean data. We test empirically the performance of different types of neural processes models on these setups, and find that the models that perform best for clean data, are different than the models that perform best under noisy conditions. Specifically, models that are based on attention, which typically outperform other models in clean conditions, are more severely affected by the presence of noise. We call this phenomena \emph{in-context overfitting}, highlighting that models with higher capacity for in-context learning, are also more susceptible to over-adapting to a noisy context. 

We then perform a more detailed study for the setup where no access to clean data is assumed, i.e. noise is present both in the training set and on the observed context at test time. We propose some adaptations to the standard training of NPs, that make them more robust to noise in this setup. These include omitting the context points from the loss term (in contrast to standard NP training methods), and controlling the weight of the predicted reconstruction variance.

We show results for both 1D functions, and image data using the CelebA dataset~\cite{liu2015faceattributes}. For both of these data types, our method outperforms all baseline models when training on noisy data. We demonstrate the importance of the two components in our method through ablations, and conclude with a discussion.

Our main contributions can be summarized as the following:
\begin{enumerate}
\item We propose to study the problem of modeling functions with noisy observations, using neural processes. As far as we know we are the first to consider this problem.
\item We show that models based on attention, which in clean conditions typically perform better than standard NP models, are more severely affected by noise, and become worse than models that do not use attention in the presence of high noise levels.
\item We propose a way to train neural processes that makes them more robust to noise. Our method is simple, easy to implement, and does not add training time.   
\end{enumerate}

All our code and data will be made available upon publication. 

\section{Preliminaries: Neural Processes} \label{sec:related_work_np}
Neural Processes (NPs) are a family of models introduced by Garnelo et al.~\cite{garneloConditionalNeuralProcesses2018, garneloNeuralProcesses2018}, with the goal of capturing the distribution of data as functions. Each data sample is represented as a function $f:\mathcal{X} \rightarrow \mathcal{Y}$, and the distribution of these functions $p(f)$, also known as a \emph{stochastic process}, is learned from a training set, using deep neural networks.  

These models are trained by sampling random points for each function, which are used as a \emph{context set} in order to predict the function and the remaining uncertainty. Following the Kolmogorov Extension Theorem~\cite{oksendal2013stochastic}, this distribution over functions is represented as a model that can output the distribution over any set of points along the function. This set of points is termed the \emph{target set}.

Since the initial models were introduced, many extensions and variations have been proposed (eg. ~\cite{kimAttentiveNeuralProcesses2019,gordonConvolutionalConditionalNeural2019,nguyen-phuocRendernetDeepConvolutional2018,rastogiSemiParametricDeepNeural2022,lee2020bootstrapping}). Many of these extensions deal with the way the context set is processed. An important set of work proposed to use attention based models and transformer architectures~\cite{vaswaniAttentionAllYou2017} to better capture the information in the context~\cite{kimAttentiveNeuralProcesses2019, nguyenTransformerNeuralProcesses2022, rastogiSemiParametricDeepNeural2022}. In contrast to standard NP models which represent the context by averaging the representation of each context point, these models have a stronger capacity to extract information by attending to different context points as a function of the position of the requested target point. This results in much better performance. Another relevant work to highlight is the BNP~\cite{lee2020bootstrapping}, which uses bootstrapping to generate multiple context sets from a single one, producing an ensemble-based prediction. 
For a comprehensive overview of NPs see Jha et al.~\cite{jha2023neural}.

Throughout the paper we compare the following models: \textbf{np}~\cite{garneloNeuralProcesses2018} - the basic NP model; \textbf{cnp}~\cite{garneloConditionalNeuralProcesses2018} - Conditional NP, a model that does not contain a latent variable and thus can only make independent predictoins of target points; \textbf{anp} and \textbf{canp}~\cite{kimAttentiveNeuralProcesses2019} - models using attention (with and without latent variables, repsectively); \textbf{banp} and \textbf{bnp}~\cite{lee2020bootstrapping} - models using bootstrapping (with and without attention, respectively). 
In our results and analysis we make a distinction between \emph{context-averaging} models: \textbf{np}, \textbf{cnp}, \textbf{bnp}, and \emph{attention-based} models: \textbf{anp}, \textbf{canp}, \textbf{banp}.
All implementations are based on the BNP code base~\cite{lee2020bootstrapping}.

\section{Related Work}

Our work is related to classical methods for signal denoising and outlier rejection (e.g. \cite{fischler1981random, buades2005non}). However, our motivation is different as our aim is to capture the distribution of an underlying signal from noisy observations rather than directly denoising the signal. Moreover, we are interested in learning the distribution from data, and study the effect of noisy training data on the standard methods to do so.

A different related topic is deep learning with noisy labels, for which multiple approaches exist.
\cite{batson_noise2self_nodate} presented a denoising method for statistically independent noise across dimensions, \cite{lehtinen_noise2noise_2018} used multiple noisy-versions of a sample to extract the clean signal, \cite{park_manner_2022} used multiple views of the data for noise erasure. 
\cite{xiao_promix_2023} has demonstrated a Sample Selection method based on calculating a sample's confidence, \cite{li_dividemix_2020} uses a mixture model and semi-supervised training to distinguish between clean and mislabeled samples, \cite{arazo_unsupervised_nodate} suggested unsupervised training of a mixture model to estimate if the sample is mislabeled. \cite{tanaka_joint_2018} proposes a joint optimization method to correct noisy labels, \cite{han_co-teaching_nodate} trains two networks at the same time to avoid confirmation bias. \cite{wei_learning_2022} defines a new benchmark called CIFAR-N, with both clean labels and noisy labels.
These works do not study the Neural Processes model, or the effect of noisy context in general. 

Most of the models we use can be viewed as a specific type of conditional variational autoencoders (VAEs)~\cite{kingmaAutoencodingVariationalBayes2014, rezendeStochasticBackpropagationApproximate2014}. Our method of tuning different terms in the loss function is related to previous work that study this for VAEs in general~\cite{higgins_-vae_2017, rezendeTamingVAEs2018}.

\section{Functions with Noisy Observations}

\begin{figure}
  \centering
  \includegraphics[width=1\textwidth]{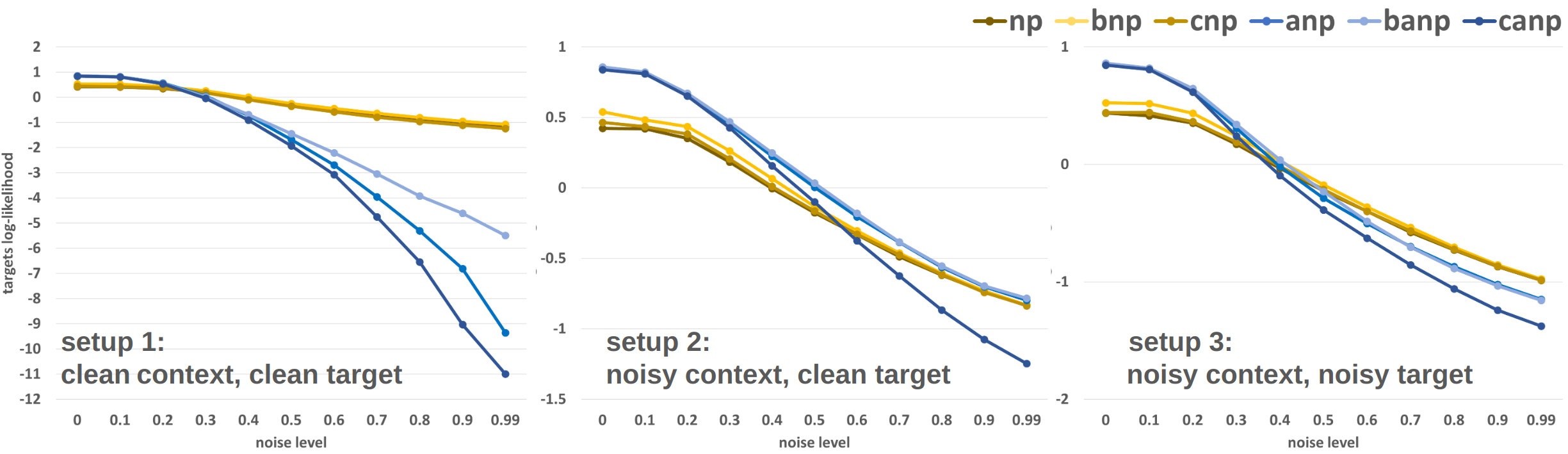}
  \caption{Prediction accuracy for three different noise setups. \textbf{Left:} When training on clean data. models perform significantly worse as the context becomes more noisy at test time. The effect is extreme in attention-based models. \textbf{Middle:} Training with noisy contexts allows the models to learn how to filter out noise and predict only the clean signal. \textbf{Right:} In this more realistic setup, models trained on noisy data cannot separate the noise from the signal. Again, attention-based models are more affected by this.}
\label{fig:noisy_training}
\end{figure}

We define the basic problem as modeling the distribution of functions, when the observed data is contaminated with noise. This means that the values of any set of points drawn from the underlying function, consist of the sum of two terms - the true value given by the function, and a noise term:
\begin{equation}
    y_{x^n} = f(x^n) + \xi^n   
\end{equation}
where $x^n$ is an arbitrary set of $n$ points in $\mathcal{X}$,  $\xi^n$ is a noise vector of dimension $n$, and we assume the dimensions of $\xi^n$ are statistically independent. For simplicity we use a notation for scalar values of $y$.

For the noise we use a simple Gaussian distribution, for which we control two parameters: 1) the standard deviation of the Gaussian noise, $s$, and 2) the noise rate $r$, which determines the fraction of points in each random set, for which the noise is added. All other points are not contaminated with noise. 
\begin{equation}
    \xi_{s, r}^n = \left\{ \begin{array}{ll} \mathcal{N}(\xi^{r \cdot n} ; 0, s^2 I) & \text{for uniformly random } r \cdot n \text{ dimensions}, 
    \\  0 & \text{for all other dimesions}\end{array} \right.
\end{equation}

In order to simplify the presentation of the results, we control these two parameters together through a single parameter $s$ which we call the \emph{noise level}. In other words, we set $r=s$, giving the same value to the standard deviation and rate of the noise. We consider the values in the range $s \in [0, 1]$, assuming $y$ in the data is normalized to a standard deviation of 1.
To view the effect of each of the parameters separately we refer the reader to the appendix. 

Given this definition of noisy observations, we are interested in three possible setups:

\textbf{setup 1}, clean context / clean target: Noise is added to the context set only at test time. This setup demonstrates the effect a noisy context has over the standard models, trained over clean data with no awareness to noise.

\textbf{setup 2}, noisy context / clean target: The same level of noise that is added to the context at test time, is added to the context at training time. This allows learning how to overcome noisy contexts. 

\textbf{setup 3}, noisy context / noisy target: All observed data is affected by noise. This means that at test time the context is noisy, and at training time both the context and target sets are noisy. Most of our work deals with this setup.

Figure~\ref{fig:noisy_training} shows the resulting prediction accuracy of different NP models in the three setups described above. In all setups we follow the standard evaluation method (as implemented in \cite{lee2020bootstrapping}), estimating the log likelihood over the target points using 50 latent samples for each function in the test set. All evaluations are done on clean target points. 

In setup 1, it is clearly shown how all models trained on clean data quickly deteriorate as the context noise level at test time increases. It is interesting to see that models that use attention (shown in different shades of blue), are affected much more severely, and even though they perform better for clean contexts, they quickly become much worse as the noise level increases. 

In setup 2, it is seen that training on noisy contexts with a loss that is computed using clean targets, allows the models to learn how to filter out the noise from the signal. In this setup, the larger capacity of attention based models enables them to perform better for all noise levels. 

In setup 3, corresponding to the more realistic setup where there is no access to any clean data, we see that the ability of the models to learn how to filter out noise is negatively affected compared to setup 2. Furthermore, attention-based models are affected more severely again. In this case, the high capacity of the models makes things worse as it allows them to capture the noise and adapt to it more than other models.

\section{Noise-Robust Training}

In this section we describe the methods we use to increase the robustness of NP models to noise. We focus on setup 3, in which we assume all training data is noisy. 

Our approach is most easily described through the loss function used for training. The general loss function used for NP models can be formulated as following~\cite{garneloNeuralProcesses2018}:
\begin{equation} \label{eq:original_loss}
    \mathcal{L}(x_\ctx, y_\ctx, x_\tar, y_\tar) = -\E_{z \sim q_\phi(z|x,y)} \log p_\theta (y | z, x, y_\ctx) 
    + \text{KL} \left( q_\phi(z|x,y) || p_\theta(z|x_\ctx,y_\ctx) \right)
\end{equation}
where $p$ and $q$ are implemented as normal distributions with means and variances modeled through deep neural networks with parameters $\theta$ and $\phi$. The prefixes \emph{\ctx} and \emph{\tar} refer to the context set and target set respectively, and $x$ and $y$ without prefixes denote the values for the union of the context and target sets. The standard way to train NP models is to include context points within the target set,  which is done to encourage the models to retain as much information about the context. Here, to make the notation explicit, we assume the context and the target are disjoint sets, and use $x$ and $y$ without prefixes to denote their union.

The main idea for making the loss robust to noise, is to decrease its dependence on $y_\ctx$ and $y_\tar$ in a way that retains the underlying global function signal as much as possible. The first way we do this is by computing the reconstruction term in the loss over target points only, excluding the context points from the prediction. This means changing the reconstruction term in Eq.~\ref{eq:original_loss} from $\log p_\theta (y | z, x, y_\ctx)$ to $\log p_\theta (y_\tar | z, x, y_\ctx)$.

While this deviates from common practice in NP training, using this reconstruction term prevents the model to directly copy points from its input to its output. The only information that can be used to predict $y_\tar$ from $y_\ctx$ is information of the underlying  function.  This update to the reconstruction term makes the training setup closer to standard models trained with noisy labels, where it can be shown that noise cancels out by the averaging of the loss on different samples. 

The reconstruction term in all NP models we consider is computed as a Gaussian distribution where the mean and variance are predicted independently for each target point $x_\tar$ conditioned on the context and latent variable:
$\mu_\theta(x_\tar, z, x_\ctx, y_\ctx), \sigma^2_\theta(x_\tar, z, x_\ctx, y_\ctx)$. 
The goal of the reconstruction variance term is to capture the pointwise stochasticity of the data.

Examples of predicted functions are shown in figure~\ref{fig:gp_plot}. The reconstruction mean for different latent samples are shown as the solid curves (30 curves per function), and reconstruction variance is depicted by the light blue area around the mean. Examining the effect of training on noisy data, as seen by comparing columns \textbf{anp} and \textbf{np} to \textbf{clean-anp} (2nd column), shows that while the pointwise variance (shaded area) increases significantly, the global variance (spread of the curves) remains low. This is not the desired effect of function prediction from a noisy context, as clearly the global uncertainty over the function needs to grow.

Following this, we propose to control the predicted variance directly in the loss function. Although the variance already appears as part of the reconstruction term, we use an additional loss term computed as the average variance for all target points. 
We also tested different ways of tuning the weight of the KL term, and other multi-objective schemes like~\cite{rezendeTamingVAEs2018}, however we find that tuning the weight of the variance loss term $\sigma^2_\theta(z, x, y_\ctx)$ alone, is enough to achieve a significant boost in test accuracy.

To summarize, we apply two updates to the loss function to make it more robust to noise: (1) We compute the loss only on the target points, excluding the context points from the reconstruction term. (2) We add the \textit{average predicted reconstruction variance} over all target points, as an additional loss term, which  penalizes large variances $\bar{\sigma}^2 
 = \frac{1}{|\tar|}\sum_{t \in \tar} \sigma^2_\theta(x_t, z, x_\ctx, y_\ctx)$. We control the term's weight through a hyper-parameter, $w_\sigma$. 

The robust loss function we use can be therefore formulated as:
\begin{eqnarray} \label{eq:proposed_loss}
    &&\mathcal{L_\text{robust}}(x_\ctx, y_\ctx, x_\tar, y_\tar) = \\ 
    && ~~~~ - \E_{z \sim q_\phi(z|x,y)} \log p_\theta (y_\tar | z, x, y_\ctx) 
    + \text{KL} \left( q_\phi(z|x,y) || p_\theta(z|x_\ctx,y_\ctx) \right) + w_\sigma \bar{\sigma}^2 \nonumber
\end{eqnarray}

\section{Results}

In the following section we describe the results of running the above modifications of the loss term on different datasets. We use various NP models as described in Sec.~\ref{sec:related_work_np}, namely \textbf{np}, \textbf{cnp}, \textbf{anp}, \textbf{bnp}, \textbf{canp} and \textbf{banp}. We add the prefix \textbf{r-} to denote that we use the robust loss term as described above. Specifically we implemented two robust models: \textbf{r-anp}, and \textbf{r-banp}, that use only the target points in the reconstruction term, and use an additional variance term. The weight of the variance term is tuned to maximize accuracy on the validation set.

\subsection{1D Gaussian Processes data}

\begin{figure}
  \centering
  \includegraphics[width=1\textwidth]{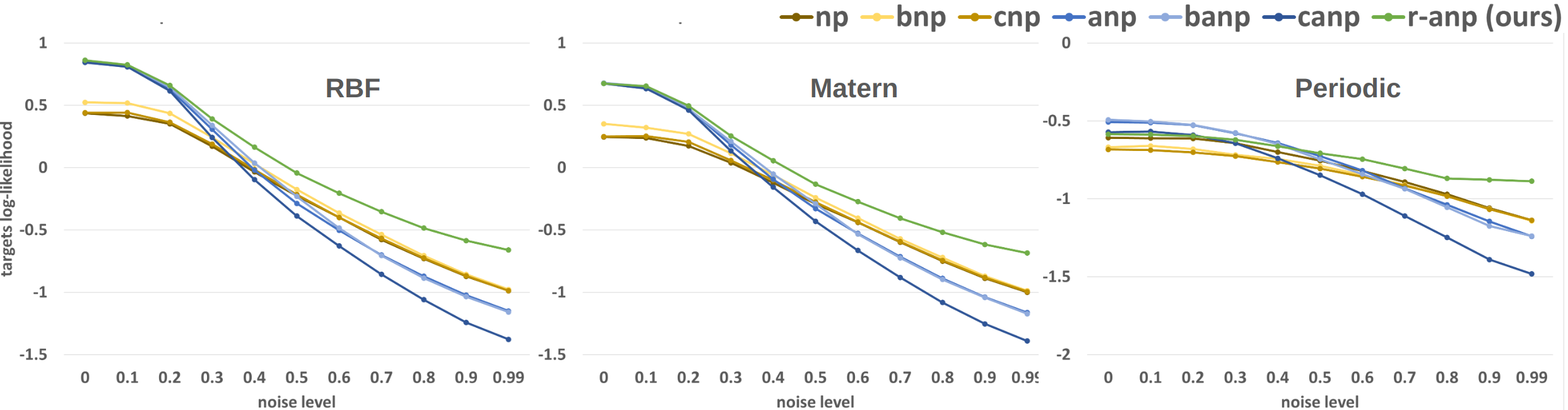}
  \caption{Target log likelihood (computed with importance sampling) for three different datasets of 1D functions. Using our robust training method on top of an attention-based model \textbf{r-anp} results in better predictions for almost all noise levels, compared to all baseline models.}
\label{fig:results_gp_kernels}
\end{figure}

\begin{figure}
  \centering
  \includegraphics[width=0.95\textwidth]{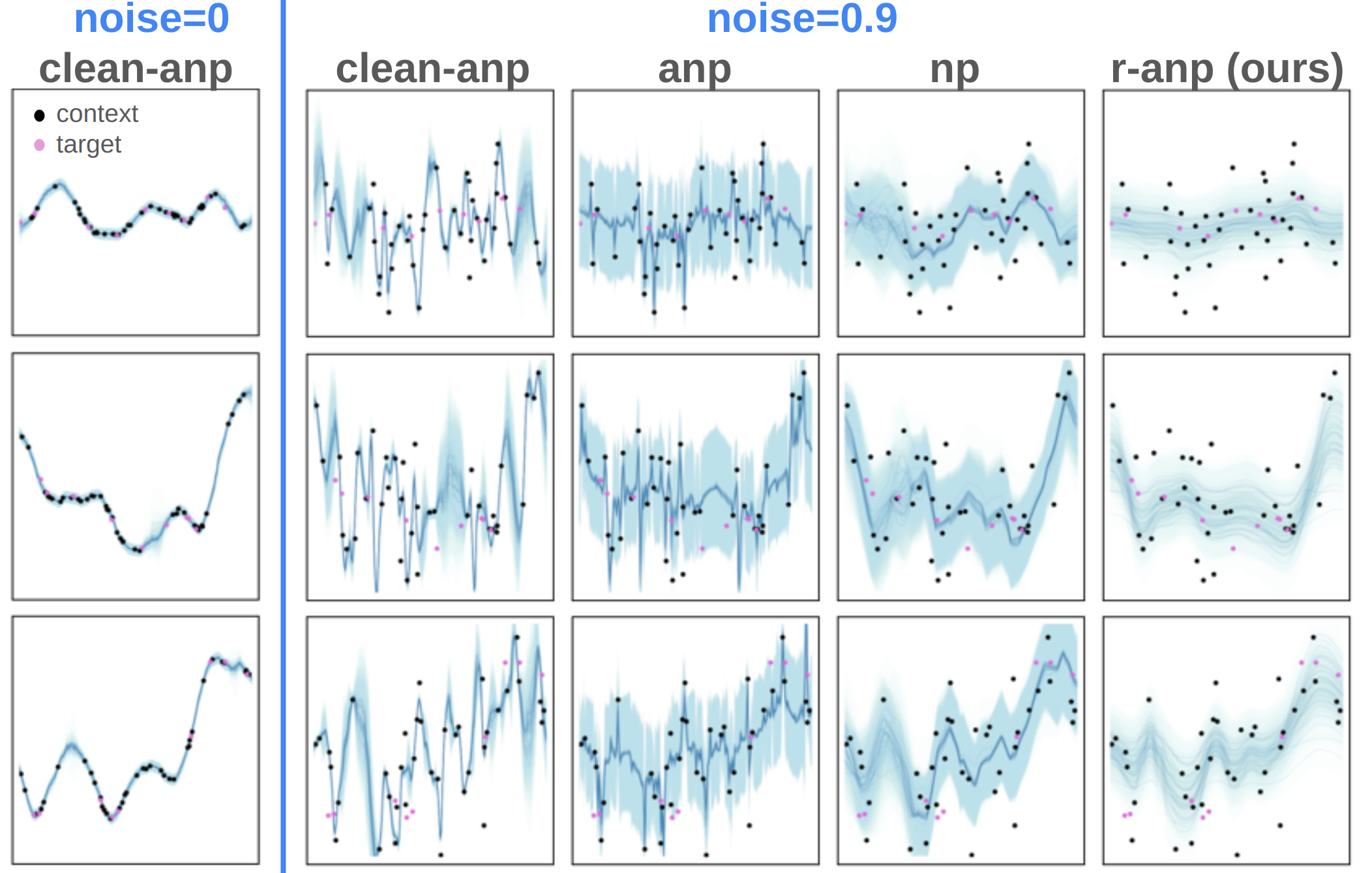}
  \caption{Examples of function predictions for models trained on Gaussian Process data with an RBF kernel. The prefix \textbf{clean-} is used for models trained on clean data. The effect of noisy context on models trained with clean data is severe in-context overfitting. Standard training with noisy data improves performance, but still results in overfitting for attention-based models (\textbf{anp}), and underfitting for context-averaging models (\textbf{np}). Our method demonstrates robust predictions with a better balance between adapting to context points, without deviating from the function prior. Furthermore, our method better captures the inherent global uncertainty as can be seen by the larger spread of the predicted function samples (the solid curves) vs. a smaller pointwise std (shaded area).}
\label{fig:gp_plot}
\end{figure}

Figure~\ref{fig:results_gp_kernels} shows the accuracy of different NP models trained on 1D functions with different noise levels. We use three different standard function families~\cite{lee2020bootstrapping}: Gaussian Process with an RBF kernel, Gaussian Process with a Matern kernel, and periodic functions. The accuracy is computed using the same method as described in Fig~\ref{fig:noisy_training} and implemented in the BNP~\cite{lee2020bootstrapping} code base. In addition to all NP models, we train a robust version of \textbf{anp} denoted by \textbf{r-anp}.

The plots show that using our robust loss function generally results in a model that is better than all other models for all noise levels. This allows attention based models to not overfit to the noise in the context, and still be more expressive than context-averaging models. We note a small region where our loss reduces accuracy, namely for periodic functions with no noise or low noise levels.

Figure~\ref{fig:gp_plot} shows the predicted function distribution of different models with two different noise levels. Each prediction consists of 30 samples of the latent variable, corresponding to a different predicted function. For each sample the thick line denotes the mean of $y$, and the shaded area the pointwise std of $y$. The left column shows the predicted function for an attention-based model trained and evaluated on clean data, and the other columns show the effect of noise at test time. A first thing to note is the effect of noise on attention-based models that were trained on clean data (\textbf{clean-anp}). The noisy context points  shift the prediction extremely around these points, but the model stays confident in its prediction, i.e. the spread of the samples, and the predicted pointwise std are small. 

A second thing to note in Fig.~\ref{fig:gp_plot} is that standard training on noisy data (\textbf{anp}, and \textbf{np}) has an effect of generally increasing the pointwise std estimate for all points. In addition, the effect is different in attention-based models and in context-averaging models: While using attention still results in strong pointwise adaptation based on the noisy context points (\textbf{anp}), context-averaging models converge to a smooth average prediction (\textbf{np}). 

Our method (\textbf{r-anp}) demonstrates both the capacity not to adapt too much to the noisy context, and at the same time maintain a distribution over the different possible underlying functions, as can be seen by the larger spread of function samples. This is in contrast to the baseline methods that increase the pointwise std to accomodate for uncertainty, and means that our model better captures the global uncertainty inherent in the problem of predicting a function from noisy observations.
We focus here on large noise levels to make the effect obvious to see. For more results using other noise levels and functions see the appendix.

\subsection{2D Images}
\begin{figure}
  \centering
  \includegraphics[width=0.85\textwidth]{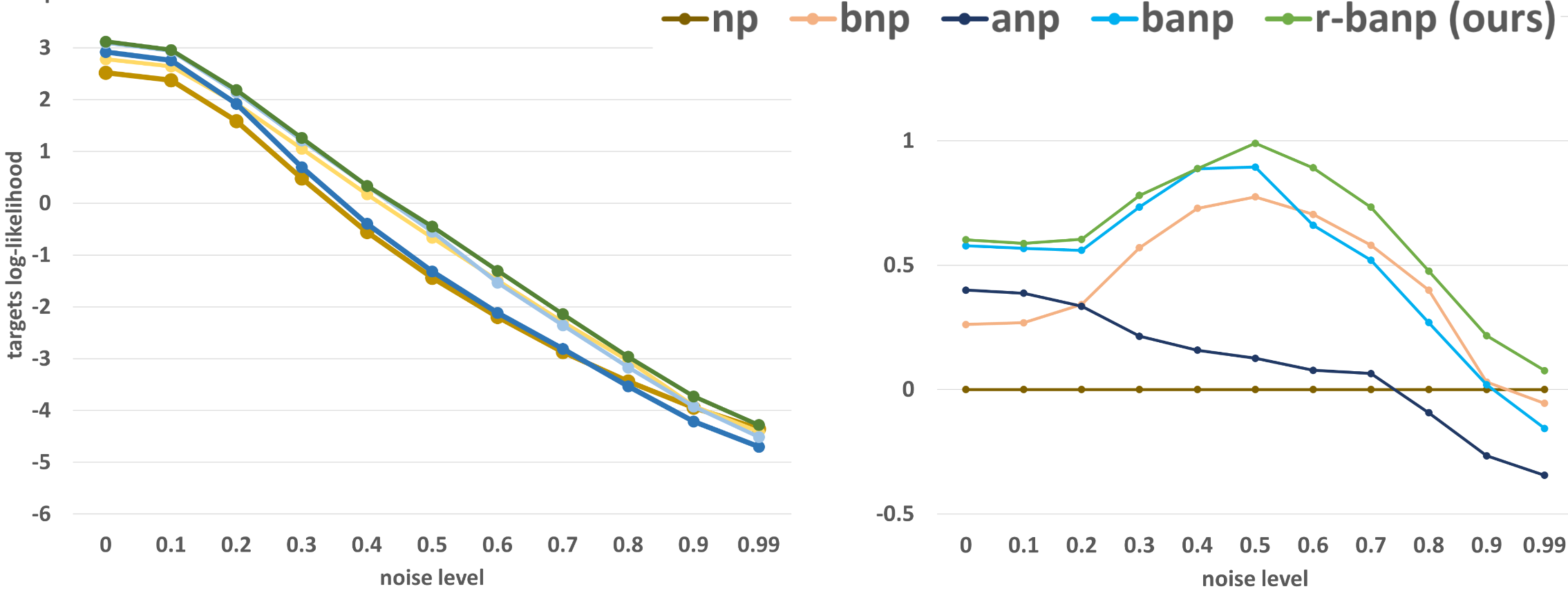}
  \caption{\textbf{Left:} Target log likelihood for the faces image dataset CelebA. Using our robust training method on top of a bootstrapping+attention based model \textbf{r-banp} results in better predictions for all noise levels, compared to all baseline models.
  \textbf{Right:} The same results showing the difference in target log-likelihood compared to the \textbf{np} model, emphasizing the gap for different noise levels.}
\label{fig:results_celeba}
\end{figure}

Following standard experiments in NP models~\cite{garneloConditionalNeuralProcesses2018,garneloNeuralProcesses2018}, we show results on 2D images, representing them as functions from pixel position to pixel color. This provides an interesting test case for NP models, because in contrast to the Gaussian Processes data used above, the distribution of images is too complex to capture in a hand-crafted manner. 

In Figure~\ref{fig:results_celeba} we show experiments on CelebA~\cite{liu2015faceattributes}, a dataset of face images. We train the same baseline models, and report the test accuracy as before. Since for this data we see a significant advantage for the boostrapping based model, we implement our robust loss on top of \textbf{banp} and denote it by \textbf{r-banp}. Our method outperforms all other models for all noise levels. 

\begin{figure}
  \centering
    ~~~~~~~~~~~~~~~~<--------------- \textbf{noise=0.6} ---------------->~~~~~~<-------------- \textbf{noise=0.9} ----------------->\\[0.05cm]
  context set = 100 pixels\\
  
  \includegraphics[trim={0cm 9.5cm 0cm 0cm},clip,width=0.99\textwidth]{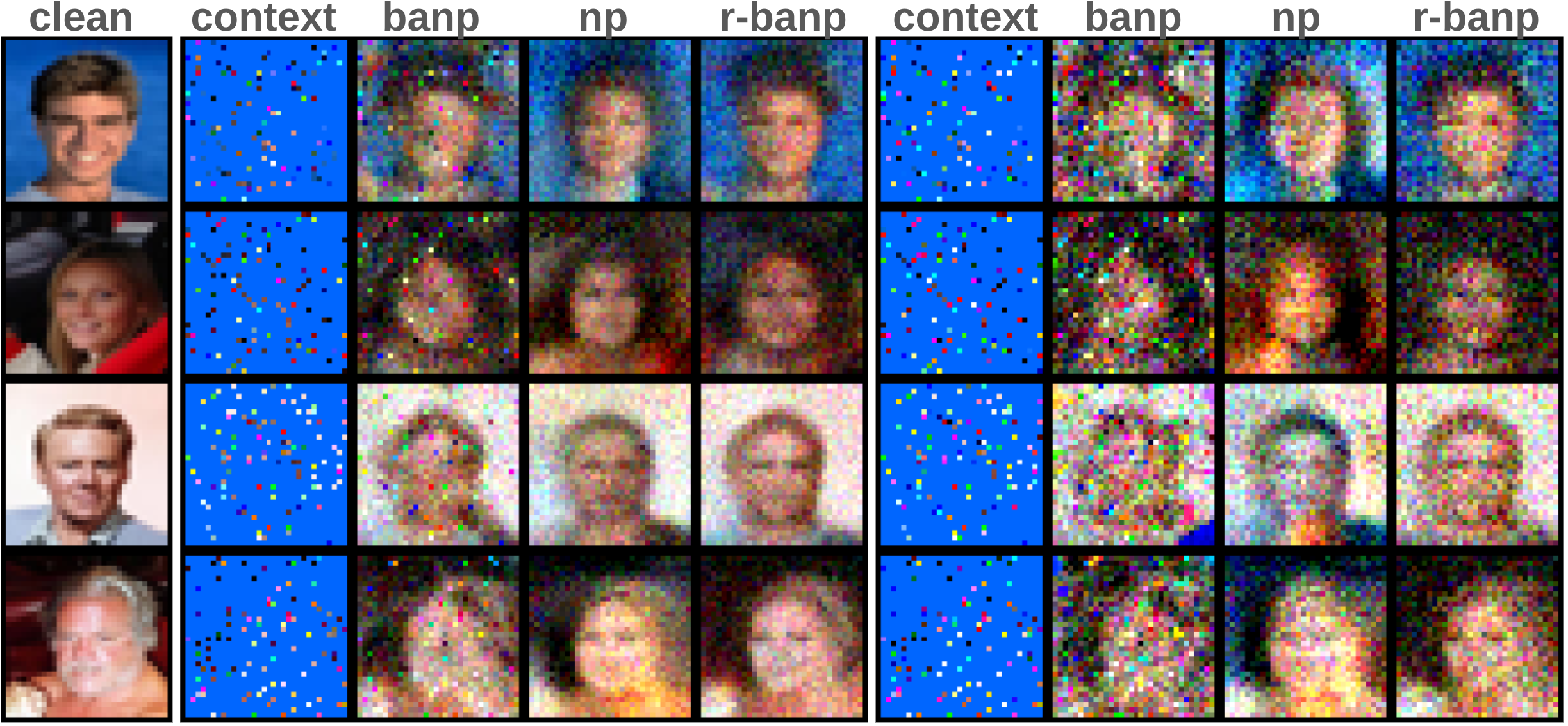} \\

  context set = 1000 pixels\\
  \includegraphics[trim={0cm 9.5cm 0cm 2cm},clip,width=0.99\textwidth]{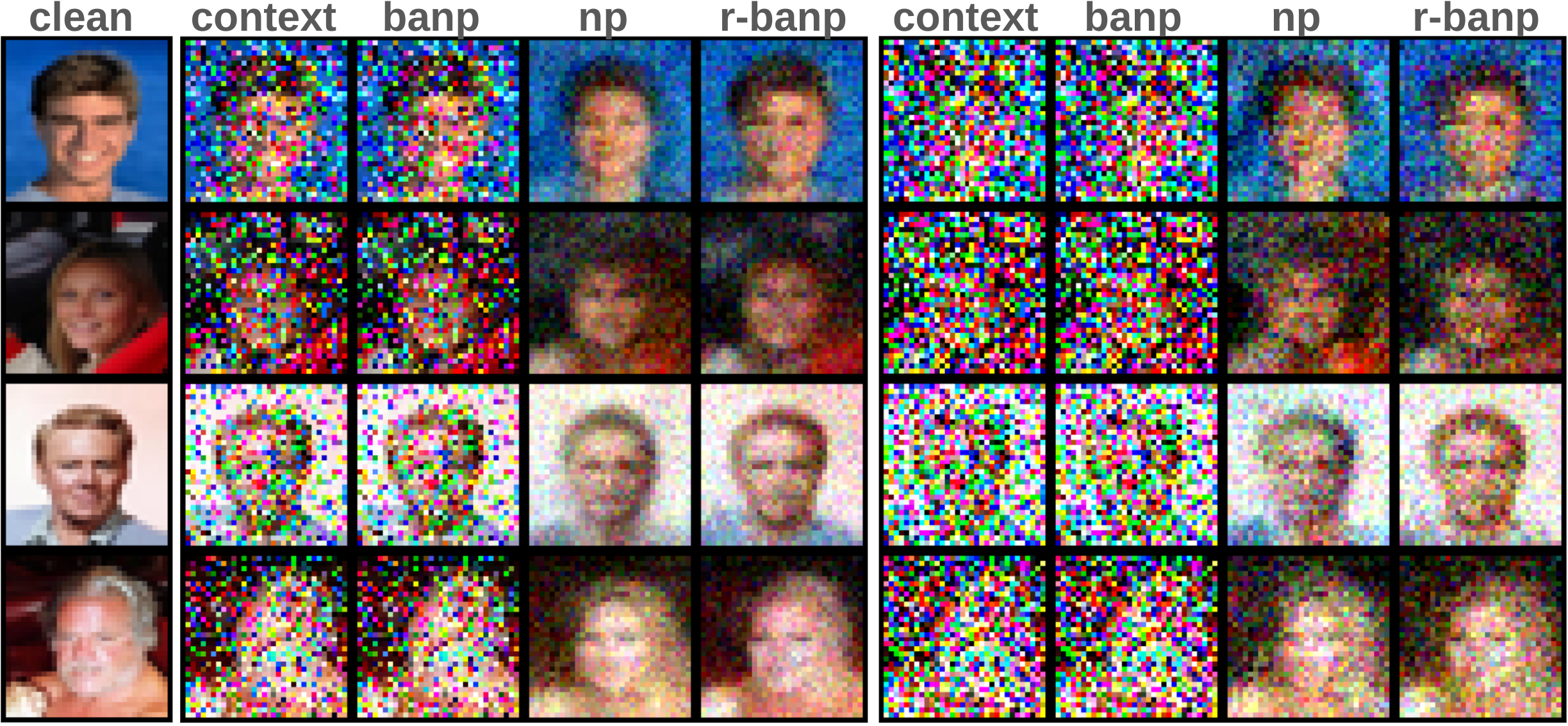}
  \caption{Examples of using NP models on image data as 2D functions. Using a context of observed pixels, the models need to predict the value for all pixels. This is especially challenging since the models never see clean images in training. Showing two different noise levels, we see that attention-based models (\textbf{banp}) overfit to the context noise, while context-averaging models (\textbf{np}) over smooth the prediction. 
  Interestingly, conditioning on a context with more points (\textbf{bottom}) can lead to worse performance, (specially evident in \textbf{banp}). Our method, \textbf{r-banp}, captures a better tradeoff between the noisy context and the prior distribution over the underlying function.} 
\label{fig:celeba_100}
\end{figure}

Examples of predicted images are shown in Fig.~\ref{fig:celeba_100}, for different noise levels and different number of context points. Each pixel in the predicted image shows the value of the mean of the predicted target point, averaging over 30 samples of the latent variable, computed by conditioning on the observed noisy context points. 
We emphasize the challenge in this task, coming from the fact that the models never see clean images in training. 
Similar to what happens in 1D functions, we see that attention-based models over-adapt to noisy context points, and context-averaging models over-smooth the prediction. Our method enables the model to better capture the inherent global uncertainty in the prediction, and leads to a better tradeoff between the observed noisy pixels, and the learned prior of images. More examples with different noise levels are shown in the appendix.

\section{Discussion \& Analysis}

\begin{figure}
  \centering
  \includegraphics[width=1\textwidth]{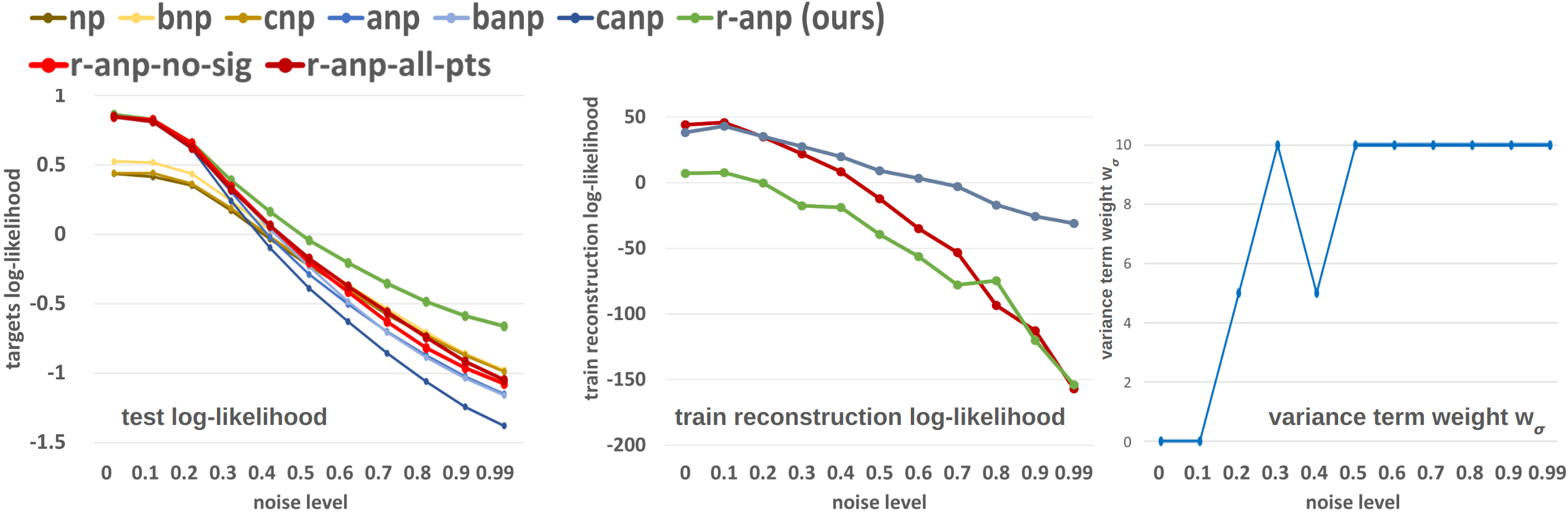}
  \caption{\textbf{Left:} Ablations of the two updates we propose. Using only one of the updates (shown in different shades of red) leads to a significant drop in performance compared to our method (green). \textbf{r-anp-no-sig} stands for not using the variance term in the loss, and \textbf{r-anp-all-pts} stands for computing the reconstruction term on all points rather than just the targets. \textbf{Middle:} The reconstruction log-likelihood term on the train set for the optimal model for each noise level (found by tuning the variance weight $w_\sigma$ to maximize validation accuracy), for \textbf{r-anp} and \textbf{r-anp-all-pts}. As the noise level increases, making the reconstruction term drop faster (compared to its trend in \textbf{anp}), leads to better accuracy at test time. \textbf{Right:} The optimal weight of the variance term $w_\sigma$ for \textbf{r-anp} at different noise levels. Tested values are $\small{[0, 5, 10, 20, 50]}$. In general stronger noise leads to higher optimal weights.}
\label{fig:ablations}
\end{figure}

We presented the problem of training NP models with noisy data, and we proposed a method that improves their robustness in this setting. Our method consists of two components: (1) training on target points only, and (2) adding a penalization term for the predicted pointwise variance. The ablation study in Fig.~\ref{fig:ablations} left shows the importance of those two components.

The updated loss function in our method, is essentially a way to control the model's reconstruction capacity, and can be viewed as a way to direct the reconstruction likelihood towards some optimal value that depends on the noise level. As seen in Fig.~\ref{fig:ablations} middle, we find that even though the reconstruction likelihood becomes naturally smaller as the noise level increases, making it even smaller is essential for improving the accuracy. Fig.~\ref{fig:ablations} right shows that the optimal value of the variance term weight $w_\sigma$ tends to increase when the noise levels become higher.

The challenges of training context based models like neural processes with noisy data, stem from the intricate relation between input and output of the model. Namely, the division between context and target, defines the noise correlations in the model's input and output. Our method can be viewed as a way to reduce the model's capacity to fit the context noise, while retaining the capacity to fit the signal - the predictive information between context and target. Although in some cases such models have already demonstrated very good results on real world data, we argue that studying their behavior in simple cases in the presence of noise, is an interesting standalone problem, that can help understand and improve such models for real world applications.

\bibliographystyle{unsrt}
\bibliography{refs}


\newpage
\appendix

\section{Additional Information}
\subsection{Limitations}
Although we tried a large range of noise levels, there are other types of noise which we haven’t tried. We focused on datasets of Gaussian Processes functions and 2D images, we haven’t tried other modalities such as Audio, Video or 3D scenes.

For the 3 GP kernels, we ran the experiments comparing all past models and RANP, for 4 times each. Each run used a different generated train set and each kernel was tested on 2 different test sets. For CelebA we ran once per configuration due to computation constraints.

\subsection{Training Details}
We based our code on the code published with the BNP model~\cite{lee2020bootstrapping}. We used a larger test set batch size of 160 samples. For optimizing our additional hyperparameter, the coefficient of the variance loss-term, we made one coarse pass on the values {-10,-1,0,1,10}, and another fine-grained pass depending on which coefficient value was the best. Finding the hyperparameter for each dataset/kernel/noise level usually didn’t take more than 2 passes, most of the times it took much less.

\subsection{Compute}
We used an NVIDIA DGX machine, each model was trained on a single A100 GPU. For the GP datasets, most models take 1 hour or less to train and BANP takes 1.5-2 hours. For the CelebA dataset most models take 4 hours to train and BANP takes 8 hours.

\section{More results}

We provide more results in the following tables and figures.

\begin{figure}
  \centering
  \includegraphics[width=0.95\textwidth]{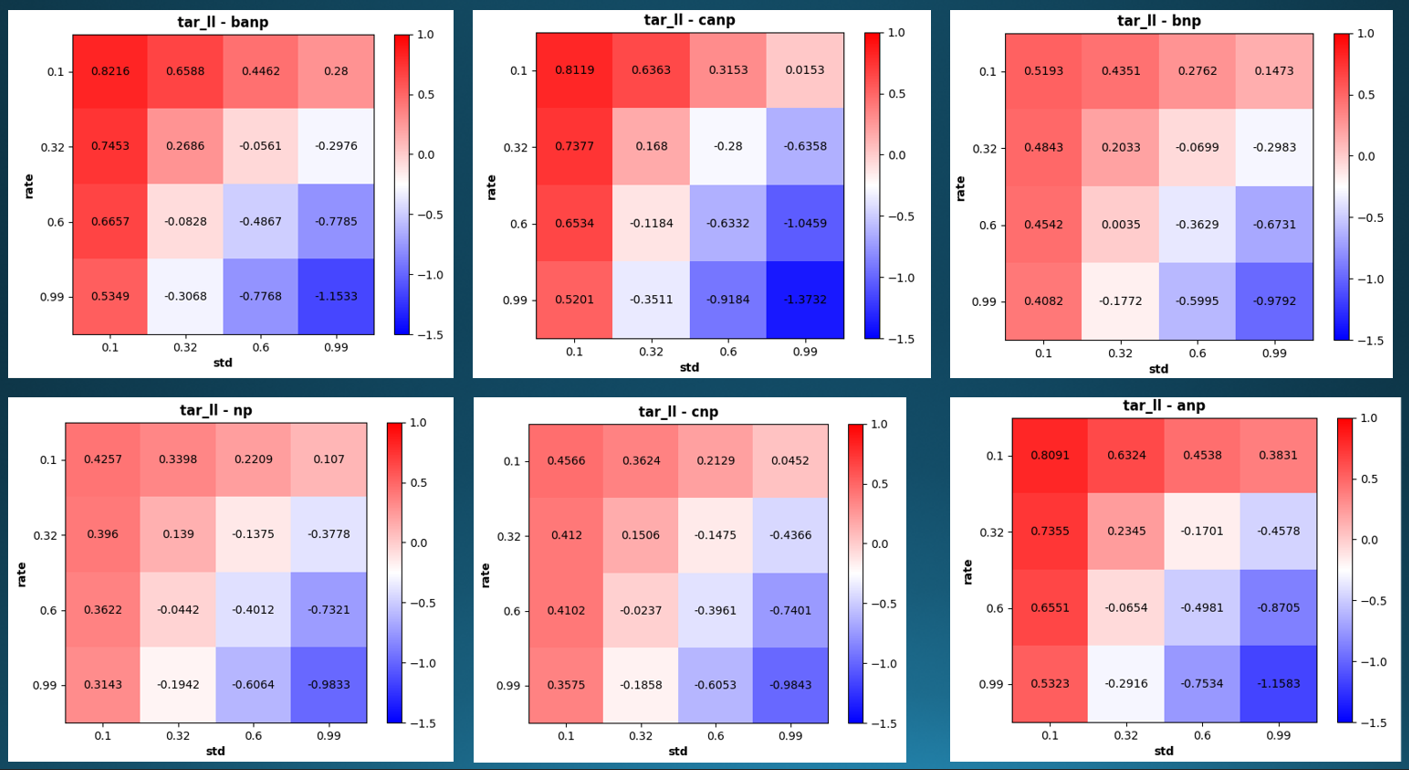}
  \caption{Results for for training different NP models on noisy data of a GP with an RBF kernel, when the noise std and noise rate are changed separately. Similar to what is shown if Fig.~\ref{fig:noisy_training}, and Fig.~\ref{fig:results_gp_kernels}, best models for low noise std and low noise rate (\textbf{anp, banp}) are different than the best models for high noise std and high noise rate (\textbf{np, bnp}).}
\label{fig:full_noise_levels}
\end{figure}

\begin{table}
    \centering
    \small{
    \begin{tabular}{|c|c|c|c|c|c|} \hline 
         kernel&  model&  noise 0&  noise 0.3&  noise 0.6& noise 0.99\\ \hline 
 & & & & &\\ \hline 
         rbf&  anp&  0.84572+-0.00309&  0.03814+-0.00687&  -2.7251+-0.02953& -8.82866+-0.44926\\ \hline 
         rbf&  banp&  0.85613+-0.00132&  0.02405+-0.00219&  -2.2207+-0.01359& -5.40225+-0.20551\\ \hline 
         rbf&  bnp&  0.52571+-0.00166&  0.25202+-0.00139&  -0.45436+-0.00621& -1.06515+-0.01457\\ \hline 
         rbf&  canp&  0.842+-0.00229&  -0.05203+-0.00927&  -3.1562+-0.06643& -10.92177+-0.3287\\ \hline 
         rbf&  cnp&  0.44818+-0.00816&  0.16955+-0.00808&  -0.58891+-0.00812& -1.23708+-0.0049\\ \hline 
         rbf&  np&  0.42242+-0.00849&  0.18659+-0.00145&  -0.51347+-0.00569& -1.15631+-0.00778\\ \hline        
     \end{tabular}     
    }    
    \caption{Numerical results for setup 1 on GP with an RBF kernel (same experiments as shown in Fig.~\ref{fig:noisy_training}) including error bars. Means and std values are computed from 3 to 4 repetitions of the experiment per configuration, with different random seeds and different test sets. }
    \label{tab:setup1}
\end{table}

\begin{table}
    \centering
    \small{
    \begin{tabular}{|c|c|c|c|c|c|} \hline 
         kernel&  model&  noise 0&  noise 0.3&  noies 0.6& noise 0.99\\ \hline 
 & & & & &\\ \hline 
         rbf&  anp&  0.84791+-0.00349&  0.45246+-0.00216&  -0.20206+-0.00768& -0.81049+-0.0103\\ \hline 
         rbf&  banp&  0.85758+-0.00055&  0.46894+-0.00104&  -0.18059+-0.00007& -0.7808+-0.00346\\ \hline 
         rbf&  bnp&  0.51928+-0.01662&  0.2609+-0.00371&  -0.30214+-0.00276& -0.83494+-0.00315\\ \hline 
         rbf&  canp&  0.84031+-0.001&  0.4253+-0.00405&  -0.3646+-0.00987& -1.22797+-0.02377\\ \hline 
         rbf&  cnp&  0.45903+-0.00413&  0.19965+-0.00374&  -0.32722+-0.00163& -0.83737+-0.00174\\ \hline 
         rbf&  np&  0.42392+-0.00843&  0.18177+-0.00437&  -0.33357+-0.00097& -0.83451+-0.0019\\ \hline 
    \end{tabular}
    }
    \caption{Numerical results for setup 2 on GP with an RBF kernel (same experiments as shown in Fig.~\ref{fig:noisy_training}) including error bars. Means and std values are computed from 3 to 4 repetitions of the experiment per configuration, with different random seeds and different test sets. }
    \label{tab:setup2}
\end{table}

\begin{table}
    \centering
    \small{
    \begin{tabular}{|c|c|c|c|c|c|} \hline 
 kernel& model& noise 0& noise 0.3& noise 0.6&noise 0.99\\ \hline 
 & & & & &\\ \hline 
         rbf&  anp&  0.8472+-0.0029&  0.3034+-0.005&  -0.5052+-0.0022& -1.1523+-0.0024\\ \hline 
         rbf&  banp&  0.8555+-0.0017&  0.3247+-0.0031&  -0.4796+-0.0016& -1.1594+-0.0034\\ \hline 
         rbf&  bnp&  0.5286+-0.0029&  0.2447+-0.0048&  -0.3634+-0.0023& -0.9787+-0.0013\\ \hline 
         rbf&  canp&  0.8433+-0.0018&  0.2492+-0.0047&  -0.6323+-0.0026& -1.3872+-0.0041\\ \hline 
         rbf&  cnp&  0.445+-0.0238&  0.1912+-0.0059&  -0.4003+-0.0016& -0.987+-0.0007\\ \hline 
         rbf&  np&  0.4295+-0.0038&  0.1697+-0.0022&  -0.4033+-0.0029& -0.9869+-0.0015\\ \hline 
         rbf&  ranp&  0.84956+-0.00708&  0.38799+-0.00486&  -0.21071+-0.01057& -0.66251+-0.00169\\ \hline 
 & & & & &\\ \hline 
 matern& anp& 0.6734+-0.0026& 0.1868+-0.0079& -0.5315+-0.0025&-1.1615+-0.0016\\ \hline 
 matern& banp& 0.6794+-0.0017& 0.2092+-0.0017& -0.5302+-0.0111&-1.1693+-0.0024\\ \hline 
 matern& bnp& 0.3348+-0.0129& 0.1144+-0.0053& -0.4066+-0.0022&-0.9874+-0.0022\\ \hline 
 matern& canp& 0.6744+-0.001& 0.1332+-0.0012& -0.6689+-0.0032&-1.3962+-0.0081\\ \hline 
 matern& cnp& 0.2437+-0.004& 0.0689+-0.0078& -0.4377+-0.0013&-0.9946+-0.0001\\ \hline 
 matern& np& 0.2462+-0.0021& 0.045+-0.0064& -0.443+-0.0039&-0.9957+-0.003\\ \hline 
 matern& ranp& 0.67241+-0.00154& 0.25055+-0.00507& -0.2774+-0.00918&-0.6845+-0.00262\\ \hline 
 & & & & &\\ \hline 
 periodic& anp& -0.5079+-0.0015& -0.5822+-0.0021& -0.8166+-0.0021&-1.2414+-0.0008\\ \hline 
 periodic& banp& -0.4934+-0.0003& -0.5771+-0.0004& -0.8404+-0.0039&-1.2596+-0.0145\\ \hline 
 periodic& bnp& -0.6648+-0.0048& -0.71+-0.0102& -0.8465+-0.0029&-1.1392+-0.0018\\ \hline 
 periodic& canp& -0.5678+-0.003& -0.6469+-0.0032& -0.9735+-0.0024&-1.4877+-0.0051\\ \hline 
 periodic& cnp& -0.6693+-0.0123& -0.7249+-0.0031& -0.8531+-0.0053&-1.1383+-0.0027\\ \hline 
 periodic& np& -0.6097+-0.0011& -0.6475+-0.0158& -0.8217+-0.0003&-1.1377+-0.0019\\ \hline 
 periodic& ranp& -0.58578+-0.00035& -0.6223+-0.00052& -0.74519+-0.00043&-0.88658+-0.00134\\ \hline
    \end{tabular}
}
    \caption{Numerical results for setup 3 with different GP kernels (same experiments as shown in Fig.~\ref{fig:results_gp_kernels}) including error bars. Means and std values are computed from 3 to 4 repetitions of the experiment per configuration, with different random seeds and different test sets. }
    \label{tab:setup3}
\end{table}

\begin{figure}[h]
  \centering
  \includegraphics[width=0.99\textwidth]{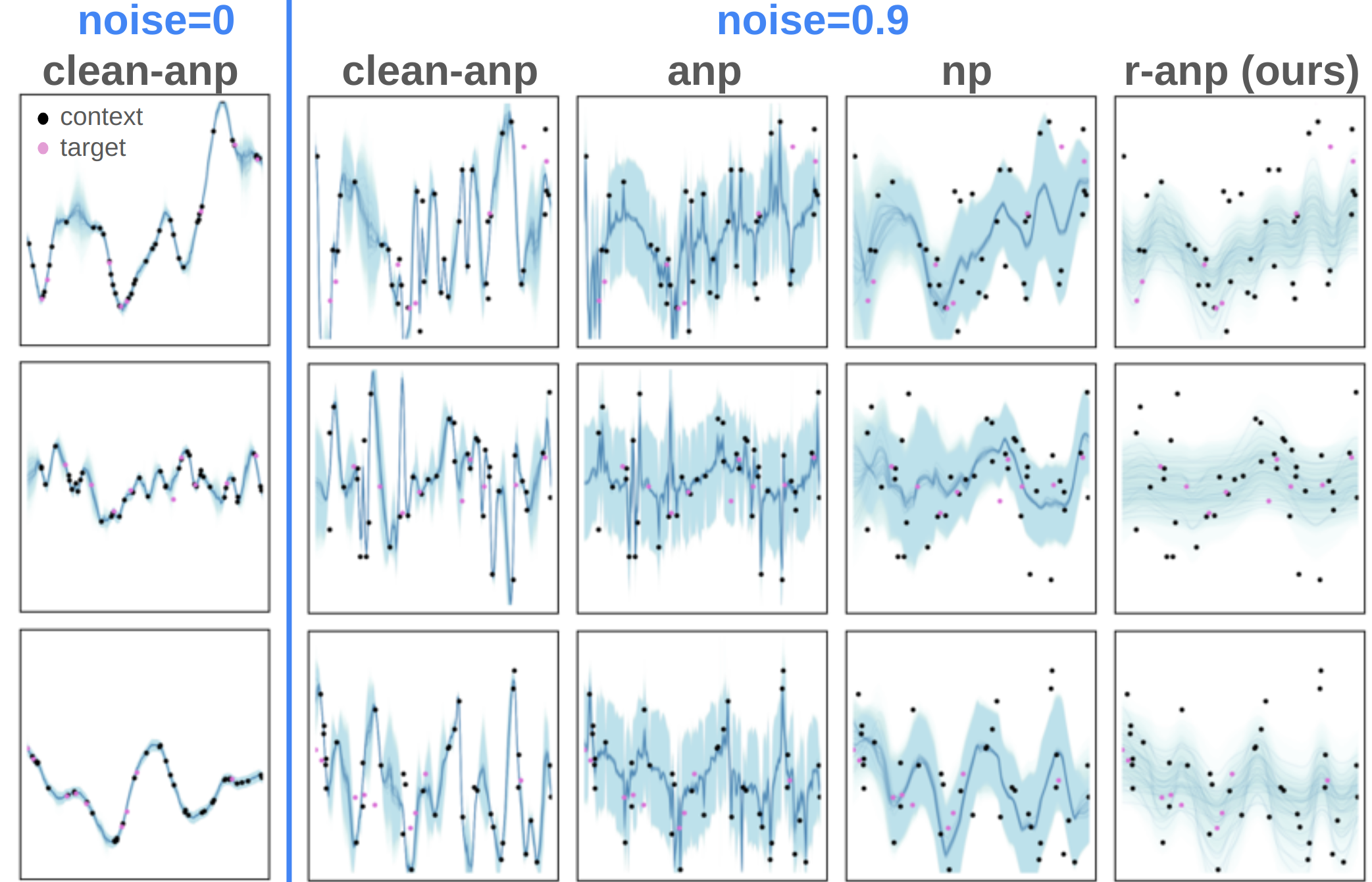}
  \caption{Examples of function predictions for models trained on Gaussian Process data with a \textbf{Matern} kernel. The prefix \textbf{clean-} is used for models trained on clean data. The effect of noisy context on models trained with clean data is severe \textit{in-context overfitting}. Standard training with noisy data improves performance, but still results in overfitting for attention-based models (\textbf{anp}), and underfitting for context-averaging models (\textbf{np}). Our method demonstrates robust predictions with a better balance between adapting to context points, without deviating from the function prior. Furthermore, our method better captures the inherent global uncertainty as can be seen by the larger spread of the predicted function samples (the solid curves) vs. a smaller pointwise std (shaded area).}
\label{fig:gp_plot_matern}
\end{figure}

\begin{figure}[h]
  \centering
  <---noise=0--->~~~~<----------------------------------noise=0.3--------------------------------->\\
  \textbf{clean-anp~~~~~~~~~~~~~clean-anp~~~~~~~~~~~~~~~~~~~anp~~~~~~~~~~~~~~~~~~~~~~~~np~~~~~~~~~~~~~~~r-anp(ours)}\\
  \includegraphics[trim={10cm 0 18cm 1.9cm},clip,width=0.99\textwidth]{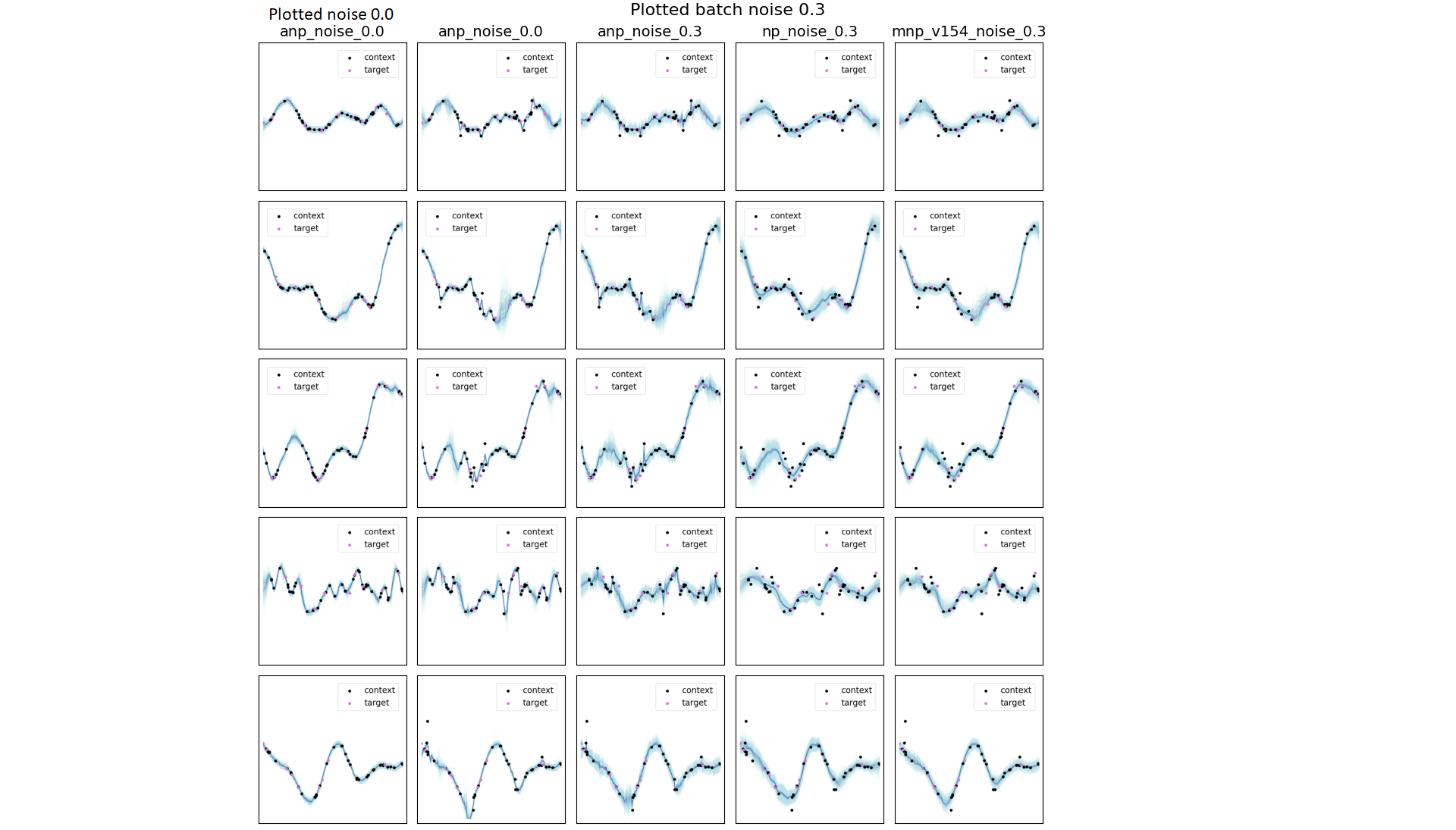}
  \caption{Examples of function predictions for models trained on Gaussian Process data with an RBF kernel and noise level 0.3.}
\label{fig:gp_plot_gp_0.3}
\end{figure}

\begin{figure}[h]
  \centering
  <---noise=0--->~~~~<----------------------------------noise=0.6--------------------------------->\\
  \textbf{clean-anp~~~~~~~~~~~~~clean-anp~~~~~~~~~~~~~~~~~~~anp~~~~~~~~~~~~~~~~~~~~~~~~np~~~~~~~~~~~~~~~r-anp(ours)}\\
  \includegraphics[trim={10cm 0 18cm 1.9cm},clip,width=0.99\textwidth]{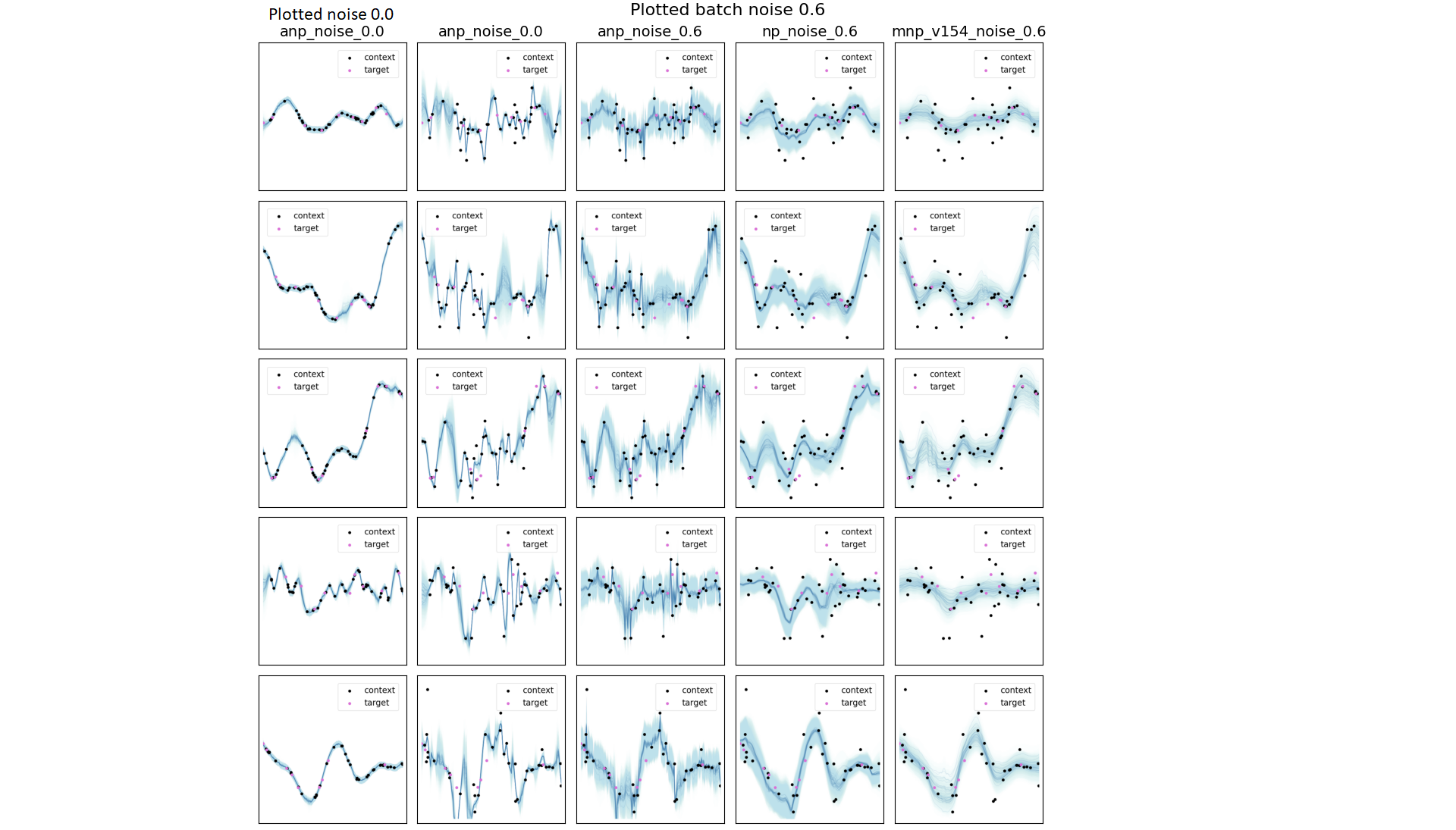}
  \caption{Examples of function predictions for models trained on Gaussian Process data with an RBF kernel and noise level 0.6.}
\label{fig:gp_plot_gp_0.6}
\end{figure}

\begin{figure}[h]
  \centering
  ~~~~~~~~~~~~~~~~<-----------------noise=0------------------->~~~~~~<---------------noise=0.3------------------>\\
  \small{\textbf{~~~~clean~~~~~~~~~~context~~~~~~~~~~~banp~~~~~~~~~~~np~~~~~~~r-banp(ours)~~~~~context~~~~~~~~~~~banp~~~~~~~~~~~np~~~~~~~r-banp(ours) }}\\
\includegraphics[trim={0cm 0 0cm 1cm},clip,width=0.55\textwidth]
{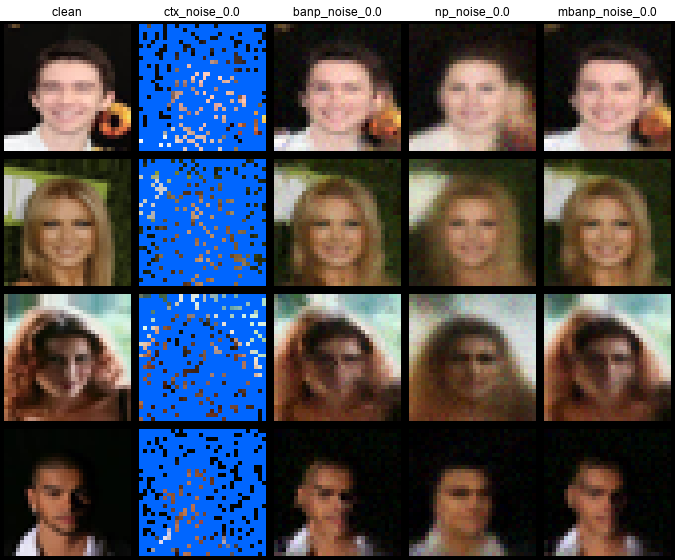}
\includegraphics[trim={4.75cm 0 0cm 1cm},clip,width=0.44\textwidth]
{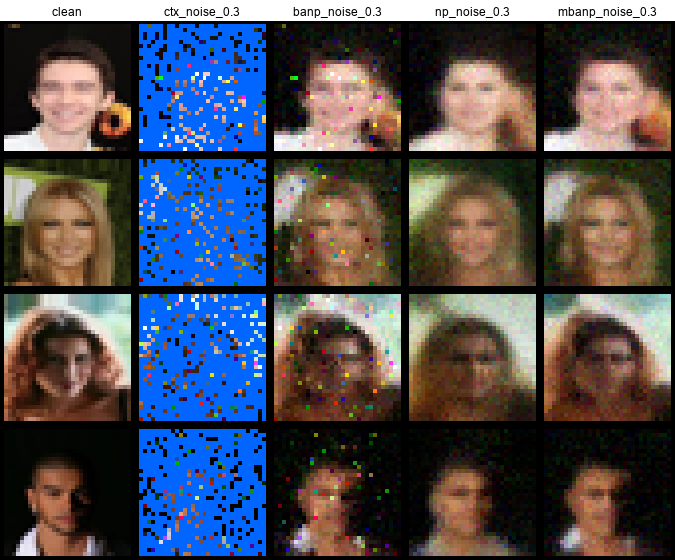} \\
\includegraphics[trim={0cm 0 0cm 1cm},clip,width=0.55\textwidth]
{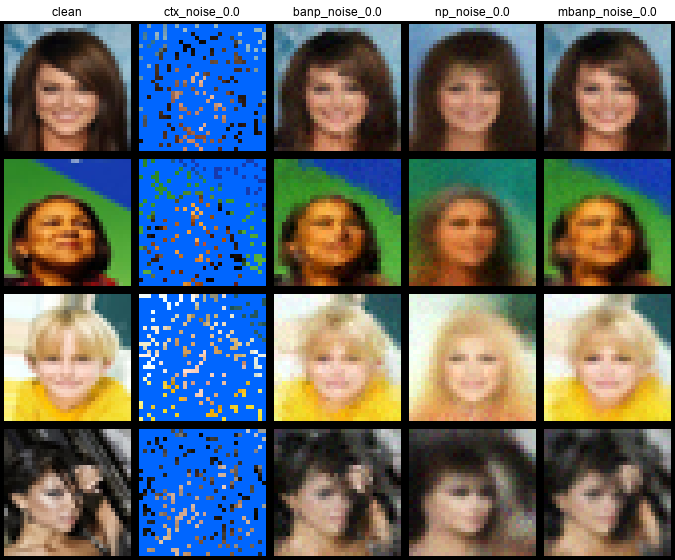}
\includegraphics[trim={4.75cm 0 0cm 1cm},clip,width=0.44\textwidth]
{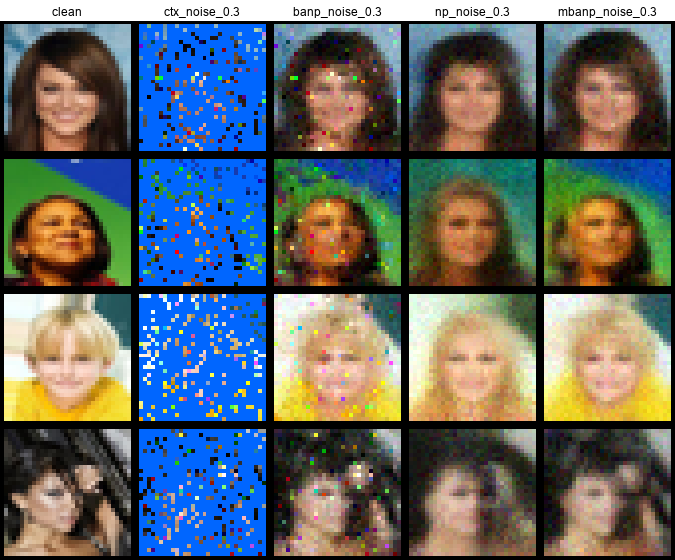} \\
\includegraphics[trim={0cm 0 0cm 1cm},clip,width=0.55\textwidth]
{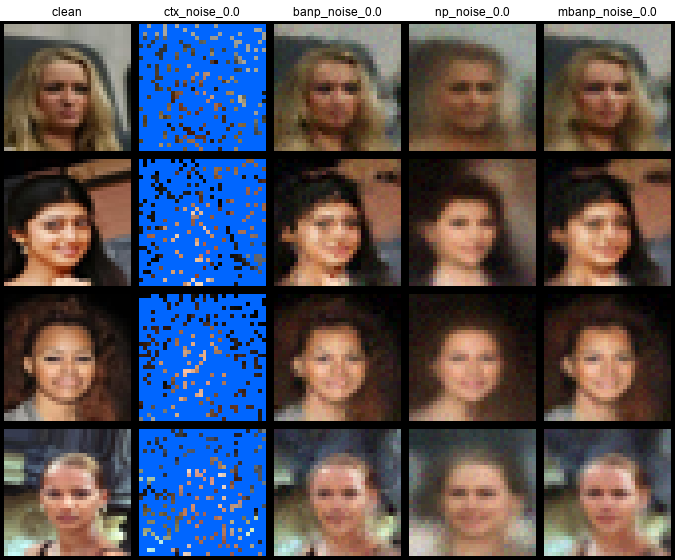}
\includegraphics[trim={4.75cm 0 0cm 1cm},clip,width=0.44\textwidth]
{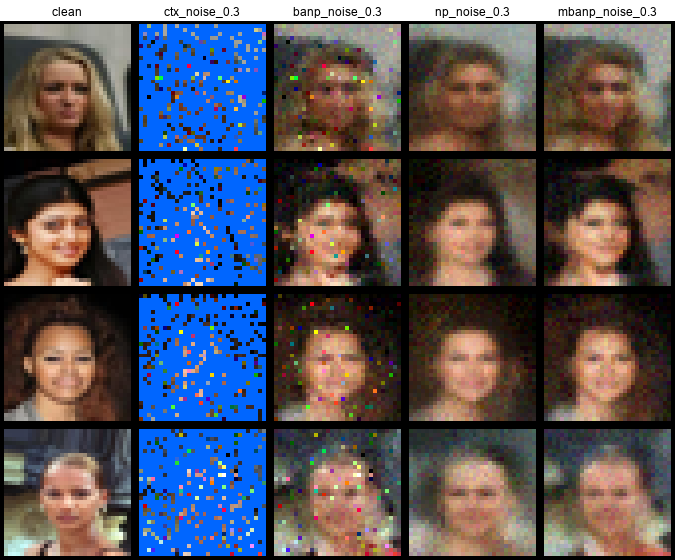} \\
  \caption{Examples of CelebA images with noisy context. We show an example with a context of 200 pixels with 0 noise on the left and a noise level of 0.3 on the right. Our method performs as good as the best baseline model for 0 noise, and outperforms other models as the noise increases.}
\label{fig:celeba_low_noise}
\end{figure}

\begin{figure}[h]
  \centering
  ~~~~~~~~~~~~~~~~<-----------------noise=0.6----------------->~~~~~~<---------------noise=0.9------------------>\\
  \small{\textbf{~~~~clean~~~~~~~~~~context~~~~~~~~~~~banp~~~~~~~~~~~np~~~~~~~r-banp(ours)~~~~~context~~~~~~~~~~~banp~~~~~~~~~~~np~~~~~~~r-banp(ours) }}\\
\includegraphics[trim={0cm 0 0cm 1cm},clip,width=0.55\textwidth]
{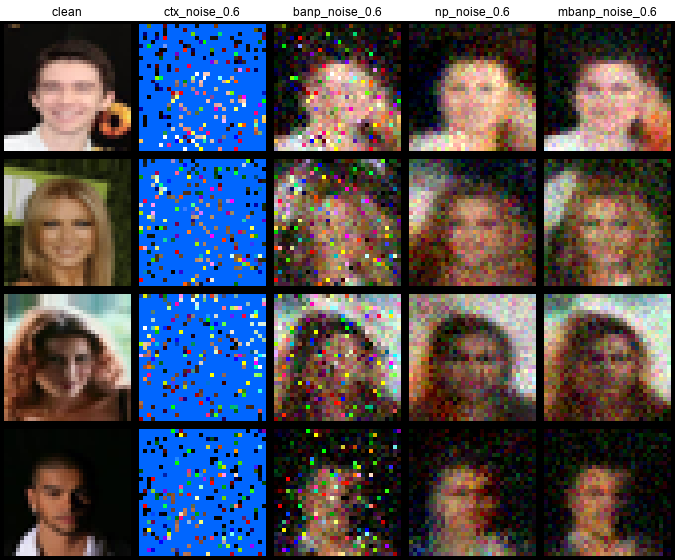}
\includegraphics[trim={4.75cm 0 0cm 1cm},clip,width=0.44\textwidth]
{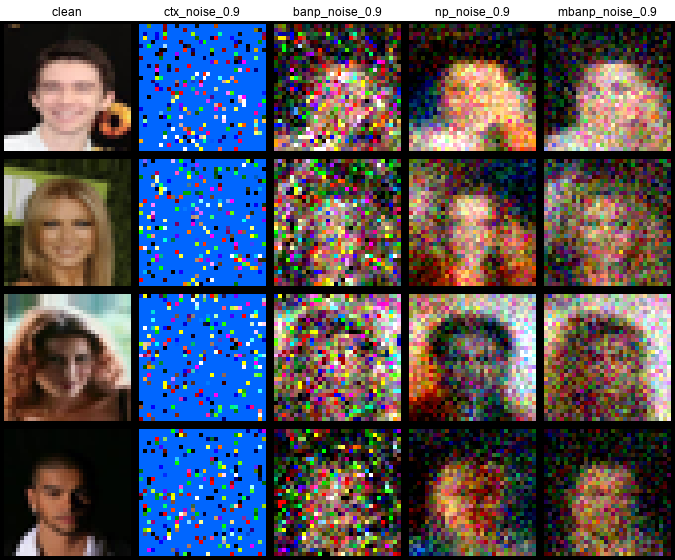} \\
\includegraphics[trim={0cm 0 0cm 1cm},clip,width=0.55\textwidth]
{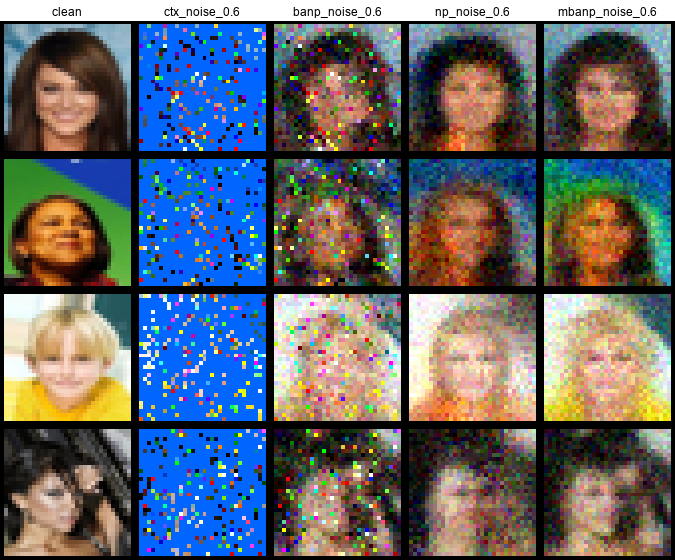}
\includegraphics[trim={4.75cm 0 0cm 1cm},clip,width=0.44\textwidth]
{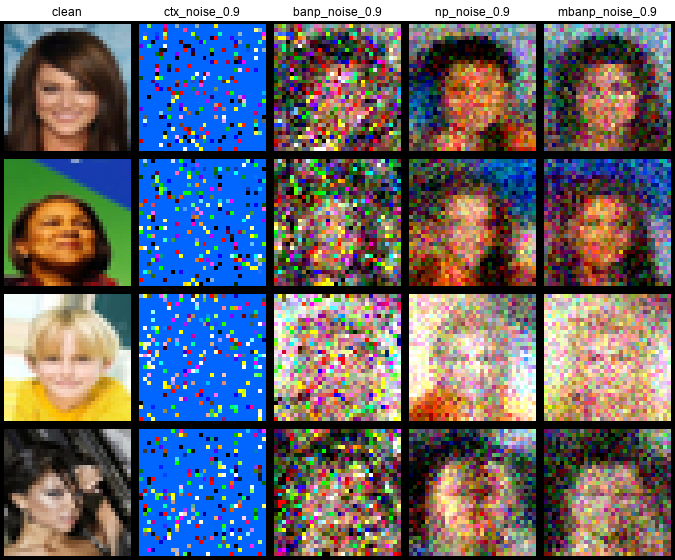} \\
\includegraphics[trim={0cm 0 0cm 1cm},clip,width=0.55\textwidth]
{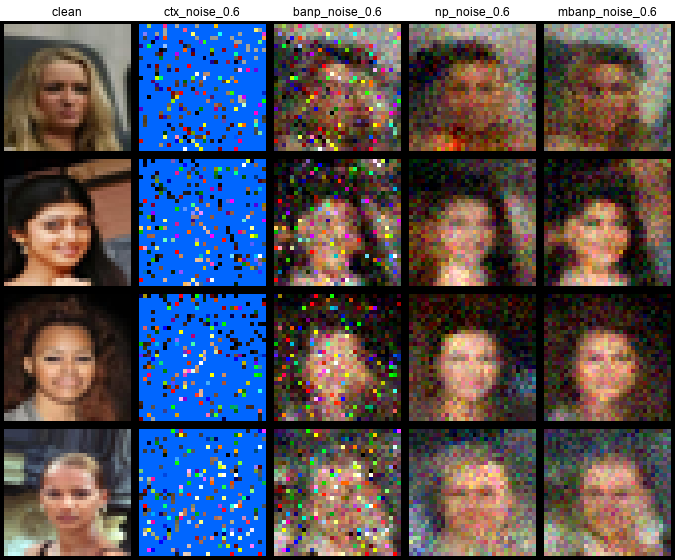}
\includegraphics[trim={4.75cm 0 0cm 1cm},clip,width=0.44\textwidth]
{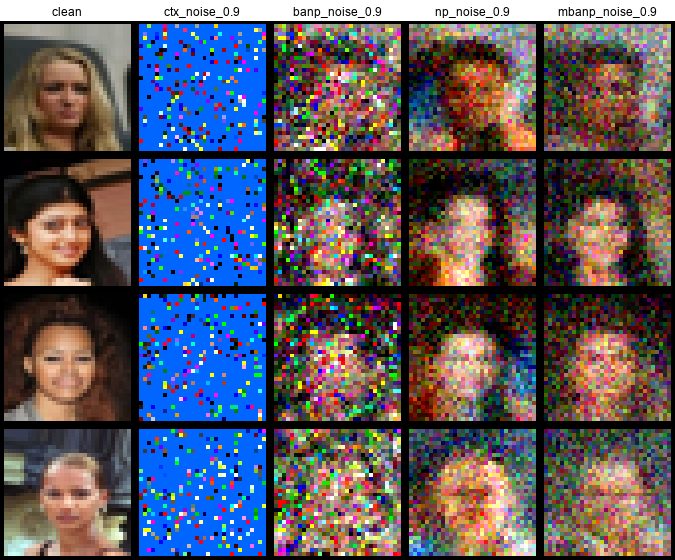} \\
  \caption{Examples of CelebA images with noisy context. We show example with a context of 200 pixels with noise level of 0.6 on the left and 0.9 on the right. Our method outperforms other models for those high noise levels. We emphasize that this is a hard problem where the models never see clean images in training.}
\label{fig:celeba_high_noise}
\end{figure}

\end{document}